\def\year{2021}\relax
\documentclass[letterpaper]{article} 
\usepackage{aaai21}  
\usepackage{times}  
\usepackage{helvet} 
\usepackage{courier}  
\usepackage[hyphens]{url}  
\usepackage{graphicx} 
\usepackage{graphicx}  
\usepackage{natbib}  
\usepackage{caption} 
\usepackage[switch]{lineno}
\usepackage{amsmath}
\title{End-to-End Learning for Simultaneously Generating Decision Map and Multi-Focus Image Fusion Result}
\author{
	Boyuan Ma, \thanks{~indicates equal contribution.}\textsuperscript{\rm 1,2,3} 
	Xiang Yin, ${}^*$\textsuperscript{\rm 1,2,3} 
	Di Wu, \textsuperscript{\rm 4}
	Xiaojuan Ban, \thanks{~corresponding authors: banxj@ustb.edu.cn.}\textsuperscript{\rm 1,2,3}\\
}
\affiliations{
	\textsuperscript{\rm 1}Institute of Artificial Intelligence, University of Science and Technology Beijing, China.\\
	\textsuperscript{\rm 2}Beijing Advanced Innovation Center for Materials Genome Engineering, University of Science and Technology Beijing, China.\\ 
	\textsuperscript{\rm 3}Beijing Key Laboratory of Knowledge Engineering for Materials Science, Beijing, China.\\
	\textsuperscript{\rm 4}Department of ICT and Natural Sciences, Norwegian University of Science and Technology, Aalesund, Norway.\\
}

\begin{document}
	\maketitle
	
	\begin{abstract}
		The general aim of multi-focus image fusion is to gather focused regions of different images to generate a unique all-in-focus fused image. Deep learning based methods become the mainstream of image fusion by virtue of its powerful feature representation ability. However, most of the existing deep learning structures failed to balance fusion quality and end-to-end implementation convenience. End-to-end decoder design often leads to unrealistic result because of its non-linear mapping mechanism. On the other hand, generating an intermediate decision map achieves better quality for the fused image, but relies on the rectification with empirical post-processing parameter choices. In this work, to handle the requirements of both output image quality and comprehensive simplicity of structure implementation, we propose a cascade network to simultaneously generate decision map and fused result with an end-to-end training procedure. It avoids the dependence on empirical post-processing methods in the inference stage. To improve the fusion quality, we introduce a gradient aware loss function to preserve gradient information in output fused image. In addition, we design a decision calibration strategy to decrease the time consumption in the application of multiple images fusion. Extensive experiments are conducted to compare with 19 different state-of-the-art multi-focus image fusion structures with 6 assessment metrics. The results prove that our designed structure can generally ameliorate the output fused image quality, while implementation efficiency increases over 30\% for multiple images fusion.
	\end{abstract}
	
	\section{Introduction}
	\label{sec:introduction}
	The multi-focus image fusion is an important topic in image processing. The limitation of optical lenses naturally presents that only objects within the depth-of-field (DOF) have a focused and clear appearance in a photograph, while other objects are likely to be blurred. Hence it is difficult for objects at varying distances to all be in focus in one camera shot~\cite{LI2017100}. Many algorithms have been designed to create an all-in-focus image by fusing multiple source images that capture the same scene with different focus points. The fused image can be used for visualization and further processing, such as object recognition and segmentation.
	
	Deep learning based solutions~\cite{lecun2015deep} are accepted to be the prevailing choice for image fusion by virtue of its powerful feature representation ability. Yu Liu introduced a convolution neural network (CNN) to image fusion and proposed a CNN-Fuse fusion method to recognize which part of the image is in-focus with a supervised deep learning structure~\cite{LIU2017191}. CNN-Fuse reached a better performance compared to traditional fusion algorithms based on the handcrafted features. Boyuan Ma moved further in applying an unsupervised training strategy to fuse images, termed as SESF-Fuse ~\cite{ma2019sesffuse}. It avoided heavy labeling work for images to train the network.
	
	Although deep learning has reached relatively good performance in multi-focus image fusion, the new problems yielded with complex structure design remain unsolved. There are three questions that deserve higher priorities. 1) The balance between fusion quality and end-to-end implementation convenience. Some structures tried to use a decoder to directly output the final fused result~\cite{Huang2020AGA,  zhang2020rethinking}. However, they did not preserve true pixel values in the source image and hardly achieve good performance in fusing evaluation due to the nonlinear mapping mechanism in the decoder. Some other structures generated intermediate decision map (DM) to reconstruct fused result with high quality~\cite{LIU2017191, xu2020deep}. But they relied highly on post-processing method (or consistency verification) choices. These methods require empirical parameters to rectify the DM, resulting in limits the generalization of them to different scenes of image fusion. 2) The gradient feature contains rich beneficial information for multi-focus image fusion. However, it was overlooked in many designs. Some deep learning structures used the l2 and SSIM objective functions to optimize the network~\cite{Li2019DenseFuseAF, Prabhakar_2017_ICCV}. These made the gradient feature completely lost during training procedure. 3) The efficiency in multiple images fusion. Currently, most of the multi-focus fusion structures focus on two images based fusion application. With multiple images fusion, the strategy is to go one by one in serial sequence~\cite{ma2019sesffuse}. However, the time consumption is scarcely acceptable for big volume image fusion.  

	In order to counterpoise the requirements of fused image quality and training simplicity, we design a gradient aware cascade structure, termed GACN\footnote{The code and data are available at \url{https://github.com/Keep-Passion/GACN}}. It simultaneously generates decision map and fused result with an end-to-end training procedure. The original pixel values in the source are retained to optimize output fused image bypassing empirical post-processing methods. Furthermore, we modify a commonly used gradient based evaluation metric as the training loss function in order to preserve gradient information. For multiple images fusion, we simplify redundant calculations by proposing a calibration module to acquire the activity levels of all images. It helps to significantly decrease the time consumption. We highlight our contributions as follows:
	
	\begin{itemize}
		\item We propose a network to simultaneously generate decision map and fused result with an end-to-end training procedure.
		\item We introduce a gradient aware loss function to preserve gradient information and improve output fusion quality.
		\item We design a decision calibration strategy for multiple images fusion in order to increase implementation efficiency.
		\item To prove the feasibility and efficiency of the proposed GACN, we conduct extensive experiments to compare with 19 different state-of-the-art (SOTA) multi-focus image fusion structures with 6 assessment metrics. We implement ablation studies additionally to test the impact of different loss function in our structure. The results prove that our designed structure can generally ameliorate the output fused image quality, and increase implementation efficiency over 30\% for multiple images fusion.
	\end{itemize}
	
	\section{Related Work}
	\label{sec:realted_work}
	The existing solutions for multi-focus image fusion can be generalized into two orientations: handcrafted feature based and deep learning based algorithms.
	\subsection{Handcrafted Feature Based Fusion Algorithms}
	\label{sec:handcrafted_feature_based_fusion_algorithms}
	Handcrafted feature based fusion algorithms concentrate on the profound image analysis of transform or spatial domains. 
	Transform domain based algorithms adopt decomposed coefficients from a selected transform domain to measure different activity levels in the input source images, such as laplacian pyramid (LP)~\cite{Burt_1983_TOC} and non-subsampled contourlet transform (NSCT)~\cite{ZHANG20091334}. 
	Spatial domain based algorithms measure activity levels with gradient features, such as spatial frequency~\cite{LI2001169}, multi-scale weighted gradient (MWG)~\cite{ZHOU201460}, and dense SIFT (DSIFT)~\cite{LIU2015139}. 
	\subsection{Deep Learning Based Fusion Algorithms}
	\label{sec:deep_learning_based_fusion_algorithms}
	Deep learning based algorithms provide prevalent solutions to image fusion problems.
	CNN-Fuse~\cite{LIU2017191} first used a convolutional network to automatically learn features in each patch of image and decided which patch is the clarity region, which achieved better performance compared to handcrafted feature based algorithms. Afterward, some researchers tried to modify the network to improve the fusion quality or efficiency. Han Tang proposed a pixel-wise fusion CNN to further improve the fusion quality~\cite{Tang2017Pixel}. Dense-Fuse~\cite{Li2019DenseFuseAF}, U2Fusion~\cite{xu2020u2fusion}, and SESF-Fuse~\cite{ma2019sesffuse} fused images in the unsupervised training procedure. IFCNN~\cite{Yu2020IFCNN} presented a general image fusion framework to handle different kinds of image fusion tasks. However, there are still other parts of deep learning based algorithm that need to refine.
	
	The output mode is an important module in network designing. Some algorithms tried to use a decoder to directly output the fused result. Hao Zhang~\cite{zhang2020rethinking} used only one convolutional layer in decoder to fuse multi-scale features and generate fused result. To improve the reconstructive ability, Hyungjoo Jung~\cite{Jung2020U} used residual block to improve the efficiency of gradient propagation, and some works~\cite{Huang2020AGA, ZHANG202140} used generative adversarial network to automatically ameliorate fusion quality. 
	However, due to nonlinear mapping in the decoder, these structures cannot precisely reconstruct fused result. This leads to relatively unrealistic performance in fusion evaluation. 
	Therefore, some structures resorted to generate an intermediate DM, to decide which pixel should appear in fused result. Some works~\cite{LIU2017191, li2020drpl} used CNN to directly output DM. SESF-Fuse~\cite{ma2019sesffuse} used spatial frequency to calculate gradient in deep features and draw out DM. Han Xu ~\cite{xu2020deep} used a binary gradient relation map to further ask decoder to preserve gradient information in DM. 
	Despite the highly fusing quality of these structures, they need some post-processing methods (or consistency verification) with empirical parameters to rectify the DM, such as morphology operations (opening and closing calculation) and small region removal strategy, which limits the generalization of the structure to different scenes of image fusion. 
	
	The objective function is a key point in structure optimization. In the field of multi-focus image fusion, the gradient in source images is an important factor to decide which part of the image is clear. 
	However, many deep learning structures only used the l2 norm and SSIM objective function to optimize the network~\cite{Li2019DenseFuseAF, Prabhakar_2017_ICCV}, which did not ask the network to preserve the gradient information in fused image. Hyungjoo Jung~\cite{Jung2020U} proposed structure tensor to preserve the overall contrast of images. Jinxing Li~\cite{li2020drpl} used an edge-preserving loss function to preserve gradient information, but it only considered gradient intensity and not took orientation information into account. In this work, we try to modify the commonly used classical gradient based evaluation metric as the loss function to directly optimize the network to export clearly fused result.
	
	Most applications of multi-focus fusion are based on multiple images. However, almost multi-focus fusion structures concentrated on two images scene and only used one by one serial fusion strategy for multiple images~\cite{LIU2017191, xu2020u2fusion}, which has in-acceptable time consumption. To the best of our knowledge, we are the first work to concentrate on the implementation efficiency in multiple images fusion scene.
	
	\begin{figure*}[t]
		\centering
		\includegraphics[width = 0.85\linewidth]{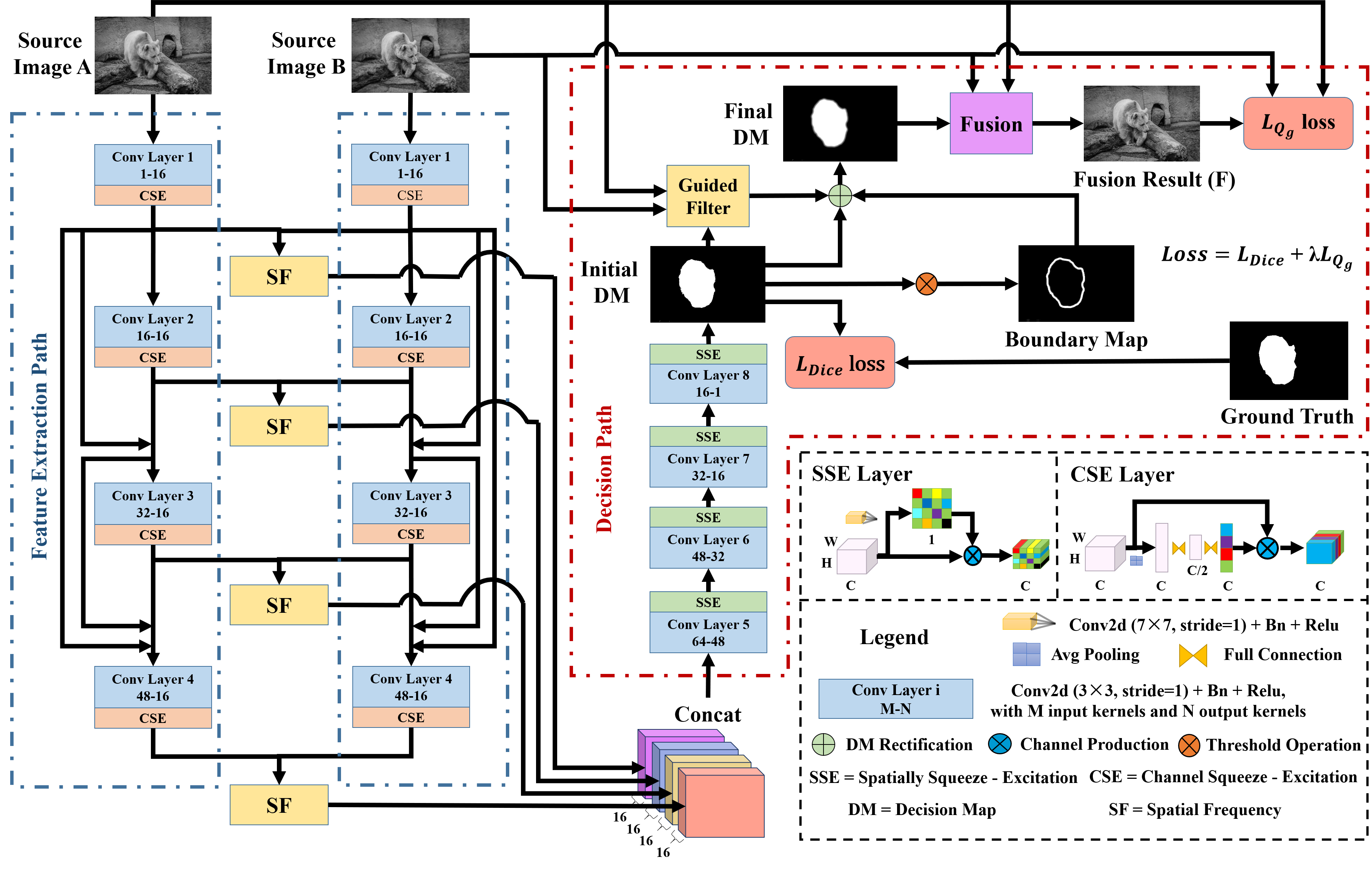}
		\caption{The network structure of the proposed algorithm.}
		\label{fig:network_structure}
	\end{figure*}
	
	\section{Method}
	\label{sec:method}
	In this section, we illustrate the details of the main contributions of this work, such as the network structure, the loss function, and the decision calibration strategy.
	\subsection{Network Structure}
	\label{sec:network_structure}
	The overall fusion network structure is shown in Figure \ref{fig:network_structure}. It includes two paths of convolutional operations, feature extraction and decision. First, we use the feature extraction path to collect multi-scale deep features of each source image. Second, we take the spatial frequency (SF) module to calculate activity level of each scale. Third, in the decision path, we concate multi-scale activity levels and feed them into some convolutional operations to draw out the initial DM, which records the probability of each pixel should be in-focused in each source image. Then we apply guided filter~\cite{6319316} to smooth the boundary of DM and acquire final DM. Finally, we cascade the fusion module in our structure and generate the fusion result.
	
	\subsubsection{Feature Extraction Path}
	\label{sec:feature_extraction_path}
	As shown in Figure~\ref{fig:network_structure}, the feature extraction path is a siamese architecture~\cite{zagoruyko2015learning}, which uses the same architectures with the same weights. It consists of a cascade of four convolutional layers to extract multi-scale deep features from each source image, and uses densely connection architecture to connect the output of each layer to the other layers, which strengthens feature propagation and reduces the number of parameters~\cite{huang2017densely, xu2020fusiondn}. To precisely localize the details of the image, there are no pooling layers in our network.
	
	In addition, we use the squeeze and excitation (SE) module after each convolutional layer, which showed good performance at image recognition and segmentation~\cite{Hu_2018_CVPR}. It can effectively enhance spatial feature encoding by adaptive recalibrating channel-wise or spatial-feature responses. Same with~\cite{ma2019sesffuse}, we use channel SE module (CSE)~\cite{roy2018concurrent} in feature extraction path. CSE uses a global average pooling layer to embed the global spatial information in a vector, which passes through two fully connected layers to acquire a new vector. This encodes the channel-wise dependencies, which can be used to recalibrate the original feature map in the channel direction.
	
	After feature extraction, we calculate multi-scale activity levels using the SF module~\cite{ma2019sesffuse}. Consider two input images $A$ and $B$, and a fused image $F$. Let $DF$ be the deep features drawn from the convolutional layer of each scale. $DF^A_{(m, n)}$ is one feature vector of pixel $i$ in source image $A$ with $(m, n)$ coordinates. We calculate its SF by:
	
	\begin{equation}
		\label{eq:rf}
		RF^A_{(m, n)}=\sqrt{\sum_{-r \leq a, b \leq r}[DF^A_{(m+a,n+b)}-DF^A_{(m+a, n+b-1)}]^2}
	\end{equation}
	\begin{equation}
		\label{eq:cf}
		CF^A_{(m, n)}=\sqrt{\sum_{-r \leq a, b \leq r}[DF^A_{(m+a,n+b)}-DF^A_{(m+a-1, n+b)}]^2}
	\end{equation}
	\begin{equation}
		\label{eq:sf}
		SF^A_{(m,n)} = \sqrt{\frac{{(CF^A_{(m,n)})}^2+{(RF^A_{(m,n)})}^2}{(2r+1)^2}}
	\end{equation}
	where $RF$ and $CF$ are respectively the row and column vector frequencies. $r$ is the kernel radius and $r=5$ in our work. The original spatial frequency is block-based, while it is pixel-based in our method. We apply the same padding strategy at the borders of feature maps.
	
	We subtract $SF^B$ from $SF^A$ to obtain activity level maps for each scale. Then we concate multi-scale activity level maps and feed them into the decision path.
	\subsubsection{Decision Path}
	\label{sec:decision_path}
	In the decision path, we first use four convolutional layers to generate the initial DM, which records the probability ($p_i$) that each pixel ($i$) of the source image $A$ is more clear than that of the source image $B$. And the initial DM is optimized by loss function with ground truth DM, as shown in the next section. 
	
	In addition, we also use the SE module in the decision path. Specifically, we use spatial squeeze and channel excitation (SSE)~\cite{roy2018concurrent}, to enhance the robustness and representatives of deep features. SSE uses a convolutional layer with one $k_s \times k_s$ kernel to acquire a projection tensor ($k_s=7$ in our work). Each unit of the projection refers to the combined representation for all channels $C$ at a spatial location and is used to spatially recalibrate the original feature map.
	
	To smooth the boundary of the fused result, we utilize threshold operation to filter the initial DM ($p_i \notin [0.1, 0.8]$) and obtain the boundary region. And then we use guided filter~\cite{6319316} to obtain the smooth DM. Finally, we use the boundary region as threshold region to combine the smooth DM and the initial DM to form the final DM. That is the boundary of the final DM is the smooth DM and the center of the final DM is the initial DM. Note that we only use a threshold operation to generate boundary region and do not hinder the backpropagation of network, which means that our structure can be trained by an end-to-end training procedure. In addition, we do not use non-differentiable post-processing methods with empirical parameters, such as morphology operation and small region removal strategy. Then, we cascade a fusion module using the final DM and source images to generate the fused result. As shown in Eq~\ref{eq:fusion}, each pixel of fused image ($F_i$) can be obtained by:
	
	\begin{equation}
		\label{eq:fusion}
		F_i = p_i \times Img^A_i + (1 - p_i) \times Img^B_i
	\end{equation}
	where the probability ($p_i$) in DM also means the fusion ratio of each pixel in the source images.
	
	Finally, we use gradient aware loss function to optimize the network to preserve gradient information in fusion result. 
	
	In general, the network can simultaneously generate DM and fusion result with end-to-end training procedure.
	\subsection{Loss Function}
	\label{sec:loss_function}
	We define a gradient aware loss function to optimize the network to simultaneously output DM and clear fusion result. The final loss function is defined in Eq~\ref{eq:loss_total}.
	
	\begin{equation}
		\label{eq:loss_total}
		L = L_{Dice} + \lambda L_{Q_g}
	\end{equation}
	where $\lambda$ is a weight to balance the importance between two losses, and $\lambda=1$ in this work.
	
	\begin{figure}[ht]
		\centering
		\includegraphics[width=1.0\linewidth]{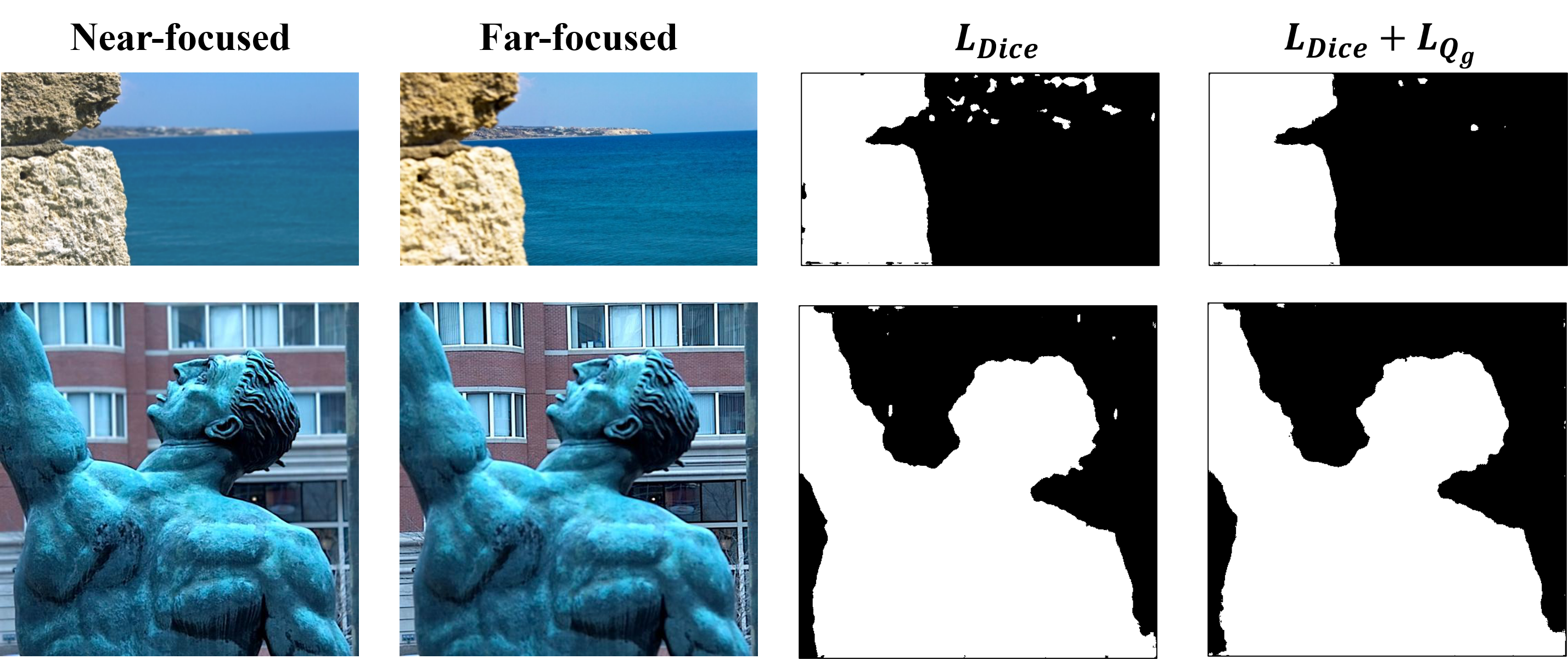}
		\caption{Visualization of decision maps of the model trained with or without $Q_g$.}
		\label{fig:vis_of_loss_qg}
	\end{figure}
	
	\begin{figure*}[ht]
		\centering
		\includegraphics[width=0.85\linewidth]{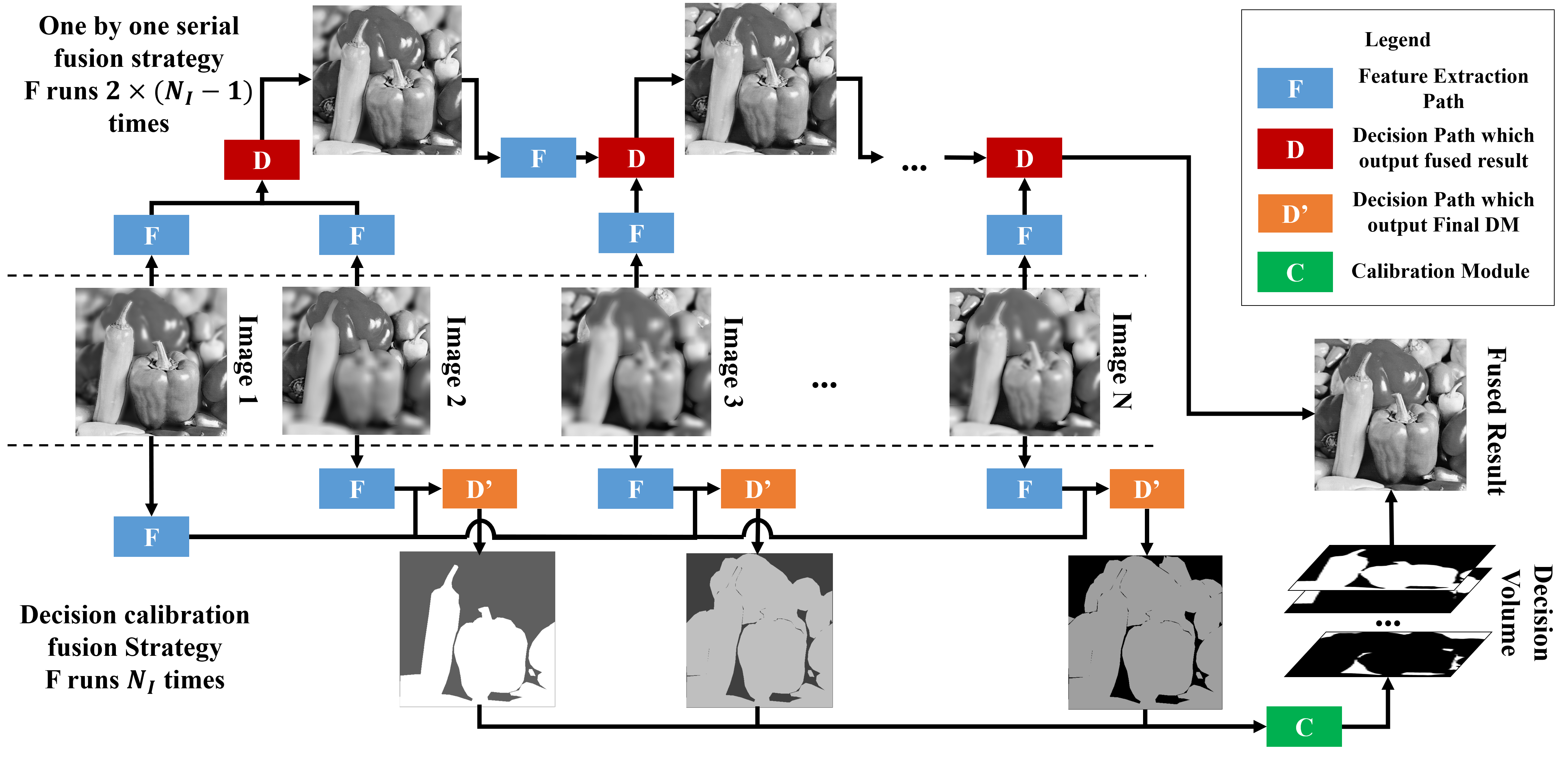}
		\caption{The flowchart of the traditional one-by-one serial fusion strategy (Top) and the proposed decision calibration strategy (Bottom).}
		\label{fig:decision_calibration}
	\end{figure*}
	
	$L_{Dice}$ is a classical loss function in semantic segmentation \cite{milletari2016vnet}, which is defined in Eq \ref{eq:loss_dice}.
	
	\begin{equation}
		\label{eq:loss_dice}
		L_{Dice} = 1 - \frac{2\sum^{N_P}_i p_ig_i + 1}{\sum^{N_P}_i p^2_i + \sum^{N_P}_ig^2_i + 1}
	\end{equation}
	where the sums run over the $N_{P}$ pixels, of the predicted binary segmentation map $p_i \in DM$ and the ground truth map $g_i \in G$. Adding 1 is to mitigate the gradient vanishing issue. 
	
	In addition, we propose to use $L_{Q_g}$ to optimize the network to export the final clear fused result. In the field of multi-focus image fusion, it is commonly speculated that only objects within the DOF have a sharp appearance in a photograph, while others are likely to be blurred~\cite{LIU2017191}. However, lots of previous works did not consider preserving gradient information in network training. In this work, we focus on a classical gradient based fusion evaluation metric, $Q_g$ or $Q^{AB}_F$~\cite{840225}, and make it differentiable as loss function in an end-to-end training procedure. By using this optimization, we lead the network to preserve gradient information in the final fused result.
	
	$Q_g$ is an evaluation metric that measures the amount of edge information transferred from input images to the fused image \cite{840225}. Consider two input images $A$ and $B$, and a fused image $F$. A sobel edge operator is applied to yield the edge strength $g_i$ and orientation $\alpha_i$ of each pixel $i$. Thus, for an input image $A$:
	
	\begin{equation}
		\label{eq:loss_qg_1}
		g{^A_i} = \sqrt{(s^{Ax}_{i})^2 + (s^{Ay}_{i})^2}
	\end{equation}
	\begin{equation}
		\label{eq:loss_qg_2}
		\alpha^A_i = tan^{-1}(\frac{(s^{Ay}_{i})^2}{(s^{Ax}_{i})^2})
	\end{equation}
	where $s^{Ax}_{i}$ and $s^{Ay}_{i}$ are the respective convoluted results with the horizontal and vertical sobel templates. 
	
	The relative strength $G^{AF}_i$ and orientation value $\Delta^{AF}_i$ between input image A and fused image F are defined as:
	
	\begin{equation}
		\label{eq:loss_qg_3}
		G^{AF}_i = 
		\begin{cases}
			\frac{g^F_i}{g^A_i}, & \text{if $g^A_i > g^F_i$} \\
			\frac{g^A_i}{g^F_i}, & \text{if $g^A_i \leq g^F_i$}
		\end{cases}
	\end{equation}
	\begin{equation}
		\label{eq:loss_qg_4}
		\Delta{^{AF}_i} = 1-\frac{| \alpha^{A}_i -\alpha^{F}_i |}{\pi/2}
	\end{equation}
	
	Unfortunately, the Heaviside function in Eq~\ref{eq:loss_qg_3} and absolute function in Eq~\ref{eq:loss_qg_4} are not differentiable and thus cannot be included in training stage. Therefore, we propose to use the sigmoid function as a smooth approximation to the Heaviside function which is defined as:
	
	\begin{equation}
		\label{eq:loss_qg_5}
		f(x, y) = \frac{1}{1 + e^{-k(x-y)}}.
	\end{equation}
	where $k$ controls the steepness of the curve and closeness to the original Heaviside function, larger $k$ means closer approximation ($k=1000$ in our work).  Then, Eq \ref{eq:loss_qg_3} can be rewritten as Eq \ref{eq:loss_qg_6}.
	
	\begin{equation}
		\label{eq:loss_qg_6}
		G{^{AF}_i} \approx f(g^F_i, g^A_i) \times \frac{g^A_i}{g^F_i} + (1 - f(g^F_i, g^A_i)) \times \frac{g^F_i}{g^A_i}
	\end{equation}
	
	And Eq \ref{eq:loss_qg_4} can be rewritten as Eq \ref{eq:loss_qg_7}.
	
	\begin{equation}
		\label{eq:loss_qg_7}
		\Delta{^{AF}_i} \approx 1-\frac{ (\alpha^{A}_i -\alpha^{F}_i) \times (2f(\alpha^{A}_i, \alpha^{F}_i) - 1)}{\pi/2}
	\end{equation}
	
	Note that in pytorch implementation \cite{paszke2019pytorch}, the gradient of absolute function is 0 when input of that equals 0, which is differentiable. Thus it can use Eq \ref{eq:loss_qg_4} rather than Eq \ref{eq:loss_qg_7} in pytorch. The detailed analysis can be found in the experiment section.
	
	The edge strength and orientation preservation values, respectively, can be derived as:
	
	\begin{equation}
		\label{eq:loss_qg_8}
		Q{^{AF}_{g_i}} = \frac{\Gamma_g}{1+e^{k_g(G^{AF}_i-\sigma_g)}}
	\end{equation}
	\begin{equation}
		\label{eq:loss_qg_9}
		Q{^{AF}_{\alpha_i}} = \frac{\Gamma_\alpha}{1+e^{k_\alpha(\Delta^{AF}_i-\sigma_\alpha)}}
	\end{equation}
	where the constants $\Gamma_g$, $k_g$, $\sigma_g$ and $\Gamma_\alpha$, $k_\alpha$, $\sigma_\alpha$ determine the shapes of the respective sigmoid functions (same with Eq \ref{eq:loss_qg_5}) used to form the edge strength and orientation preservation value. Normally, $\Gamma_g = \Gamma_a = 1$, $k_g = -10$, $k_\alpha = -20$, $\sigma_g = 0.5$, $\sigma_\alpha = 0.75$. The edge information preservation value is then defined as:
	
	\begin{equation}
		\label{eq:loss_qg_10}
		Q^{AF}_i = Q^{AF}_{g_i} \times Q^{AF}_{\alpha_i}
	\end{equation}
	
	The final assessment is obtained from the weighted average of the edge information preservation values:
	
	\begin{equation}
		\label{eq:loss_qg_11}
		L_{Q_g} = 1 - Q_g = 1 - \frac{\sum_i^{N_P}(Q{^{AF}_i}w{^{A}_i}+Q{^{BF}_i}w{^{B}_i})}{\sum_i^{N_P}(w{^{A}_i}+w{^{B}_i})}
	\end{equation}
	where $w^{A}_i = [g^{A}_i]^\gamma$ and $w^{B}_i = [g^{B}_i]^\gamma$. $\gamma$ is a constant, and usually sets $\gamma$ = 1.
	
	In total, we modify a gradient based classical fusion evaluation metric ($Q_g$) as a loss function to optimize the network to export clearly fused result. 
	
	We further show an experiment to visualize the comparison of $L_{Dice} + L_{Q_g}$ and $L_{Q_g}$, as shown in Figure~\ref{fig:vis_of_loss_qg}. It is shown that the fusion model trained with $L_{Q_g}$ have less noise in the decision map compared to the model without it, which means $L_{Q_g}$ can act as a post-processing method to improve the fusion quality because it can preserve gradient information in the image.
	
	\begin{table*}[ht]
		\centering
		\caption{Comparison with traditional methods in testing set. The bold value denotes best performance in each metric.}
		\label{table:compare_with_traditional_methods}
		\begin{tabular*}{\hsize}{@{}@{\extracolsep{\fill}}cccccccc@{}}
			\hline
			Methods                  & $Q_g$           & $Q_y$            & $Q_{ncie}$      & $Q_{cb}$        & $FMI\_EDGE$       & $FMI\_DCT$      & Time(s)\\ \hline
			GACN                     & \textbf{0.7169} & \textbf{0.97769} & \textbf{0.8411} & 0.7948          & \textbf{0.897806} & 0.4058          & 0.16\\
			MFF-GAN (2021)           & 0.5623          & 0.88652          & 0.8210          & 0.6437          & 0.884512          & 0.3699          & 0.33\\
			FusionDN (2020)          & 0.5216          & 0.82352          & 0.8209          & 0.6106          & 0.878785          & 0.3050          & 0.49\\
			U2Fusion (2020)          & 0.5590          & 0.86993          & 0.8210          & 0.6388          & 0.882694          & 0.3118          & 0.75\\
			IFCNN (2020)             & 0.6486          & 0.93751          & 0.8265          & 0.7158          & 0.891569          & 0.3757          & \textbf{0.06}\\
			PMGI (2020)     & 0.4803          & 0.80668          & 0.8209          & 0.5805          & 0.880374          & 0.3527          & 0.21\\
			SESF-Fuse (2019)      & 0.7150          & 0.97761          & 0.8397          & \textbf{0.7965} & 0.897133          & 0.3953          & 0.30\\
			Dense-Fuse (2019)  & 0.5329          & 0.83965          & 0.8239          & 0.6109          & 0.886998          & 0.4046          & 0.38\\
			CNN-Fuse (2017)           & 0.7153          & 0.97706          & 0.8396          & 0.7676          & 0.897800          & 0.4079          & 188.16\\
			DSIFT (2015)              & 0.5419          & 0.84643          & 0.8255          & 0.6306          & 0.889215          & 0.3900          & 49.28\\
			MWG (2014)             & 0.7041          & 0.97720          & 0.8376          & 0.7878          & 0.898504          & 0.3965          & 24.99\\
			Focus-Stack (2013)        & 0.5098          & 0.78907          & 0.8276          & 0.6628          & 0.868776          & 0.2332          & 0.19 \\
			SR (2010)              & 0.6792          & 0.95132          & 0.8326          & 0.7523          & 0.896763          & 0.3924          & 698.44\\
			NSCT (2009)            & 0.6721          & 0.94886          & 0.8272          & 0.7326          & 0.896647          & 0.4037          & 19.99\\
			CVT (2007)            & 0.6373          & 0.93765          & 0.8252          & 0.7111          & 0.895890          & 0.4055          & 14.76\\
			DTCWT (2007)            & 0.6688          & 0.95190          & 0.8267          & 0.7304          & 0.896893          & 0.4031          & 12.21\\
			SF (2001)                  & 0.5202          & 0.82904          & 0.8239          & 0.6173          & 0.889395          & \textbf{0.4145} & 2.25 \\
			DWT (1995)                 & 0.6444          & 0.91346          & 0.8326          & 0.6997          & 0.890219          & 0.3293          & 11.51\\
			RP (1989)                & 0.6652          & 0.94001          & 0.8280          & 0.7330          & 0.892010          & 0.3574          & 11.34\\
			LP (1983)              & 0.6834          & 0.95369          & 0.8286          & 0.7509          & 0.897242          & 0.3911          & 11.58\\ \hline
		\end{tabular*}
	\end{table*}
	
	\subsection{Decision calibration for multiple images fusion}
	
	Most applications of multi-focus fusion are based on multiple images. However, currently almost multi-focus fusion structures concentrated on two images scene and only used one by one serial fusion strategy for multiple images fusion. As shown at the top of Figure~\ref{fig:decision_calibration}, one-by-one serial strategy needs to run $2 \times (N_I-1)$ times feature extraction paths and $N_I-1$ times decision paths, where $N_I$ is the number of the source images. In this work, we propose a decision calibration strategy, which shown at the bottom of Figure~\ref{fig:decision_calibration}. It only needs to run $N_I$ times feature extraction paths and $N_I-1$ times decision paths by using the calibration module, which can generally decrease time consumption.
	
	In the decision calibration strategy, the first image is used as baseline, and feeds it to the structure with other images. Thus, we can save the parameters of the first image in the feature extraction path and avoid repeating computation. Then it uses final DMs drawn from each decision path to calculate the decision volume (DV), which records the activity levels of all the source images. The calculation process is acting as normalization to draw out relative clarity of each source images, which is shown below:
	
	
	\begin{equation}
		\label{eq:decision_calibration}
		DV^j_i = 
		\begin{cases}
			p^2_i, & \text{if $j=1$} \\
			1 - p^2_i, & \text{if $j=2$} \\
			\frac{p^2_i \times (1 - p^j_i)}{p^j_i}, & \text{$others$}
		\end{cases}
	\end{equation}
	where $j \in \{1,..,N_I\}$, is the index of the source image and $p^j_i$ is the value of pixel $i$ in final DM when fuses the source image $1$ and the source image $j$.
	
	Then, we choose the index of maximum in $DV^j_i$ for each pixel $i$ as the index of the most clarity pixel $i$ in the source images. According to the above indices, we can obtain the entire resulting fusion image.
	
	It is important to notice that the decision calibration strategy can only applied to the DM based network structure without the empirical post-processing methods. Because those empirical post-processing methods, such as morphology operation and small region removal strategy, which used in CNN-Fuse~\cite{LIU2017191} and SESF-Fuse~\cite{ma2019sesffuse}, firstly require to convert the initial DM to the binary DM, which loss the relative clarity information and can not be used in the process of decision volume calculation. Our method, GACN, can draw out the decision map without the empirical post-processing methods, which is more suited to the decision calibration strategy in the application of multiple images fusion.

	\section{Experiment}
	\label{sec:experiment}
	\subsection{Dataset}
	\label{sec:dataset_generation}
	\subsubsection{Training set}
	\label{sec:training_set}
\begin{figure}[!t]
		\centering
		\includegraphics[width=1.0\linewidth]{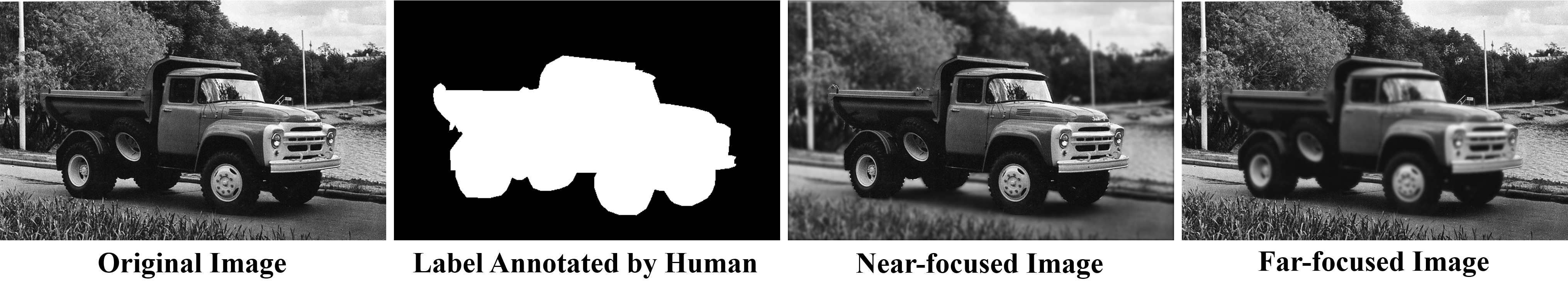}
		\caption{Visualization of a training example generated by MS COCO dataset.}
		\label{fig:training_example}
	\end{figure}

	In this paper, we generate multi-focus image data based on MS COCO dataset~\cite{lin2014microsoft}. The MS COCO dataset contains annotations for instance segmentation, and our method uses the original image and its segmentation annotation to generate multi-focus image data. That is, we use annotation as threshold region to decide which part of the image should be filtered by gaussian blurring. As shown in Figure~\ref{fig:training_example}, the original image 'truck' and its annotation are obtained by MS COCO. We use gaussian filter to blur the background to form near-focused image and blur the foreground form far-focused image. And we use the defocused spread effect model proposed in~\cite{ma2019matte} to further improve the realness of the generated data. Thus we have two inputs of multi-focus images, one ground truth fused result (original image) and one decision map (label) for network training. Because some data in MS COCO dataset contains multiple instances that are not at the same depth-of-field (DOF), so we only select images that contain one instance. Besides, we regard the multi-focus image fusion problem as an image segmentation problem. The imbalance of the foreground and background category often affects the segmentation results, so we further select the image with the foreground size between 20,000 and 170,000 pixels as the training data. Finally, we obtain 5786 images and divide these into training set and validation set according to the ratio of 7:3.
	
	\subsubsection{Testing set}
	\label{sec:testing_set}
	We use 26 image pairs of publicly available multi-focus images from \cite{NEJATI201572,savic2012multifocus} as the testing set for evaluation.
	
	\subsection{Training Procedure}
	\label{sec:training procedure}
	During training, all images were transformed to gray-scale and resized to $256 \times 256$, then random cropped to $156 \times 156$. Note that images were gray-scale in the training phase, while images for testing can be gray-scale or color images with RGB channels. For color images that needed to be fused, we transformed the images to gray-scale and calculated a decision map to fuse them. In addition, we used random crop, random blur, random offset, and gaussian noise as data augmentation methods. The initial learning rate was $1 \times 10^{-4}$, and this was decreased by a factor of 0.8 at every two epochs. We optimized the objective function by Adam \cite{kingma2015adam}. The batch size and number of epochs were 16 and 50, respectively. Our implementation was derived from the publicly available Pytorch framework \cite{paszke2019pytorch}. The network’s training and testing were performed on a station using an NVIDIA Tesla V100 GPU with 32 GB memory.
	
	\subsection{Evaluation Metrics}
	\label{sec:evaluation_metrics}
	We use six classical fusion metrics: $Q_g$~\cite{840225}, $Q_y$~\cite{2008A}, $Q_{ncie}$~\cite{wang2008performance}, $Q_{cb}$~\cite{CHEN20091421}, $FMI\_EDGE$ and $FMI\_DCT$~\cite{HAGHIGHAT2011744} to evaluate the quality of fused result.
	$Q_g$ evaluates the amount of edge information transferred from input images to the fused image. $Q_y$ calculates the similarity between fused image and the sources\cite{2008A}. $Q_{ncie}$ measures the nonlinear correlation information entropy between the input images and the fused image \cite{wang2008performance}. $Q_{cb}$ is a perceptual quality measure for image fusion, which employs the major features in a human visual system model \cite{CHEN20091421}. $FMI\_EDGE$ and $FMI\_DCT$ calculates the mutual information of the edge features and discrete cosine transform feature between the input images and the fused image \cite{HAGHIGHAT2011744}. 
	A larger value of any of the above six metrics indicates better fusion performance. For fair comparison, we use appropriate default parameters for these metrics, and all codes are derived from their public resources~\cite{imageFusionMetrics, haghighat2014fast}.
	
	\subsection{Comparison}
	\label{sec:comparison}
	To demonstrate the performance of our method, we compare it with recent SOTA fusion methods in objective and subjective assessments.
	
	\begin{figure}[ht]
		\centering
		\includegraphics[width=1.0\linewidth]{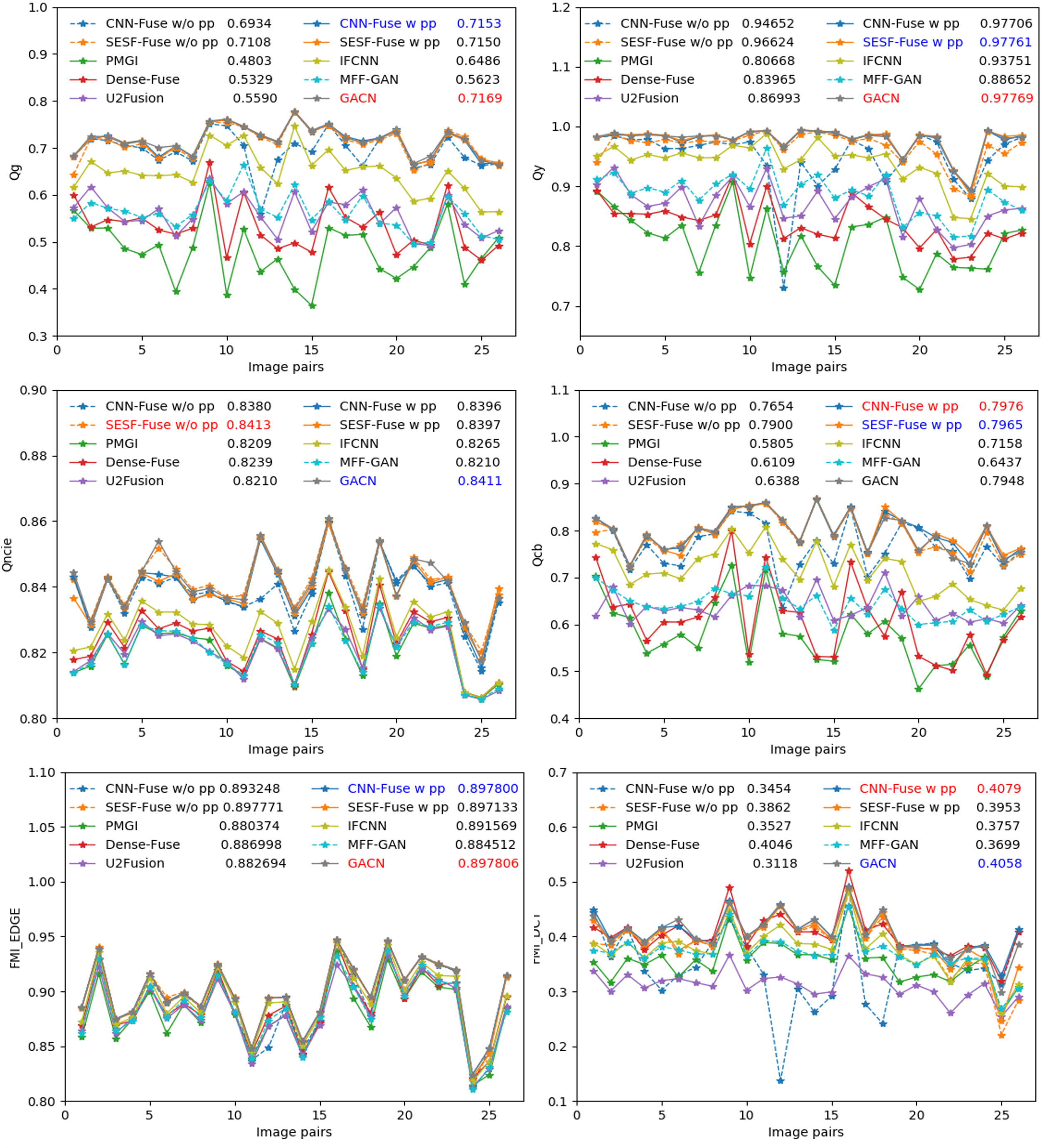} 
		\caption{Objective Assessments of our GACN with other SOTA algorithms. Means of metrics for different algorithms are shown in the legends, and evaluation for each image pair is shown in the plot. Optimal values are shown in red and sub-optimal values in blue. ‘pp’ means post processing methods.}
		\label{fig:objective_plot}
	\end{figure}
	
	\subsubsection{Objective Assessment}
	\label{sec:objective_assessments}
	
	The comparison of our method with existing multi-focus fusion methods are listed in Table~\ref{table:compare_with_traditional_methods}, such as MFF-GAN~\cite{ZHANG202140}, FusionDN~\cite{xu2020fusiondn}, U2Fusion~\cite{xu2020u2fusion}, IFCNN~\cite{Yu2020IFCNN}, PMGI~\cite{zhang2020rethinking}, SESF-Fuse~\cite{ma2019sesffuse}, Dense-Fuse~\cite{Li2019DenseFuseAF}, CNN-Fuse~\cite{LIU2017191}, dense SIFT (DSIFT) ~\cite{LIU2015139}, multi-scale weighted gradient (MWG) ~\cite{ZHOU201460}, Focus-Stack~\cite{FocusStack}, sparse representation (SR)~\cite{Yang_2010_TIM}, non-subsampled contourlet transform (NSCT)~\cite{ZHANG20091334}, curvelet transform (CVT)~\cite{NENCINI2007143}, dual-tree complex wavelet transform (DTCWT)~\cite{LEWIS2007119}, spatial frequency (SF)~\cite{LI2001169}, discrete wavelet transform (DWT)~\cite{LI1995235},  ratio of low-pass pyramid (RP)~\cite{TOET1989245}, and Laplacian pyramid (LP)~\cite{Burt_1983_TOC}. Specifically, we further show detailed comparison of each image pair with nine SOTA deep learning based methods in Figure~\ref{fig:objective_plot}. With two of them are DM based methods (CNN-Fuse and SESF-Fuse), and six of them are decoder based methods (MFF-GAN, FusionDN, U2Fusion, DenseFuse, IFCNN and PMGI). In addition, for CNN-Fuse and SESF-Fuse, we also compare different versions of whether to use post-processing (pp) methods (or consistency verification) with empirical parameters. According to experiment, DM based algorithms generate an intermediate decision map to decide which pixel should appear in the fused result, which can precisely preserve true pixel values of the source image. And decoder based algorithms directly use a decoder to draw out the fused result and cannot preserve true pixel values because of the nonlinear mapping mechanism in the decoder. Therefore, DM based algorithms achieve high performance in objective assessments, while decoder based algorithms show unrealistic performance. In addition, DM based algorithms rely on post-processing methods to rectify DM, so the performance will degrade if we remove it. Our algorithm, GACN can simultaneously generate decision map and fused result with end-to-end training procedure, and gradient information can be preserved by the gradient loss function.  Our method, achieves robust promising performance compared to above traditional methods. 
	
	In addition, the run times of different fusion methods per image pair on the test set are listed in Table~\ref{table:compare_with_traditional_methods}. Such methods as GACN, MFF-GAN, FusionDN, U2Fusion, IFCNN, PMGI, SESF-Fuse, CNN-Fuse, and DenseFuse are tested on a GTX 1080Ti GPU, and others on an E5-2620 CPU. GACN achieves an average running time of 0.16 second, which is faster than most of the methods and can be applied to actual application. Although the IFCNN is faster than GACN, it achieves lower fusion quality compared to GACN. 
	
	\begin{figure*}[!t]
		\centering
		\includegraphics[width=1.0\linewidth]{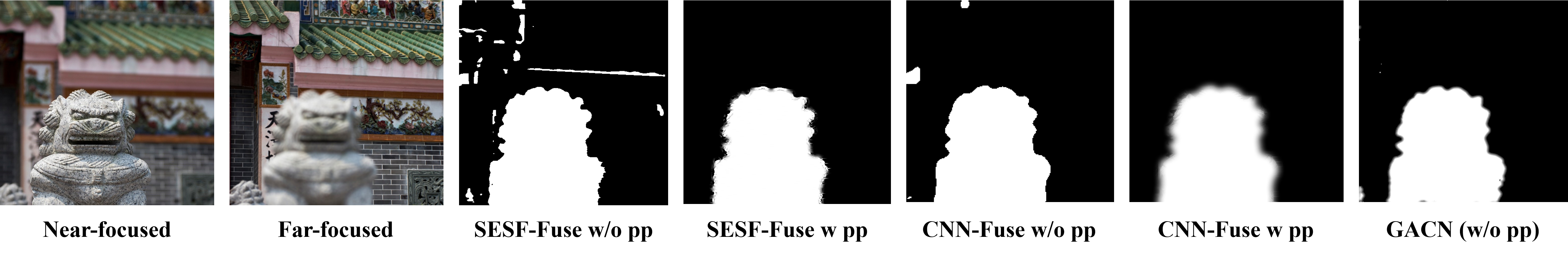} 
		\caption{Visualization of decision map of DM based methods (SESF-Fuse and CNN-Fuse) and GACN. pp means post processing methods.}
		\label{fig:vis_compare_dm}
	\end{figure*}
	
	\begin{figure*}[!t]
		\centering
		\includegraphics[width=1.0\linewidth]{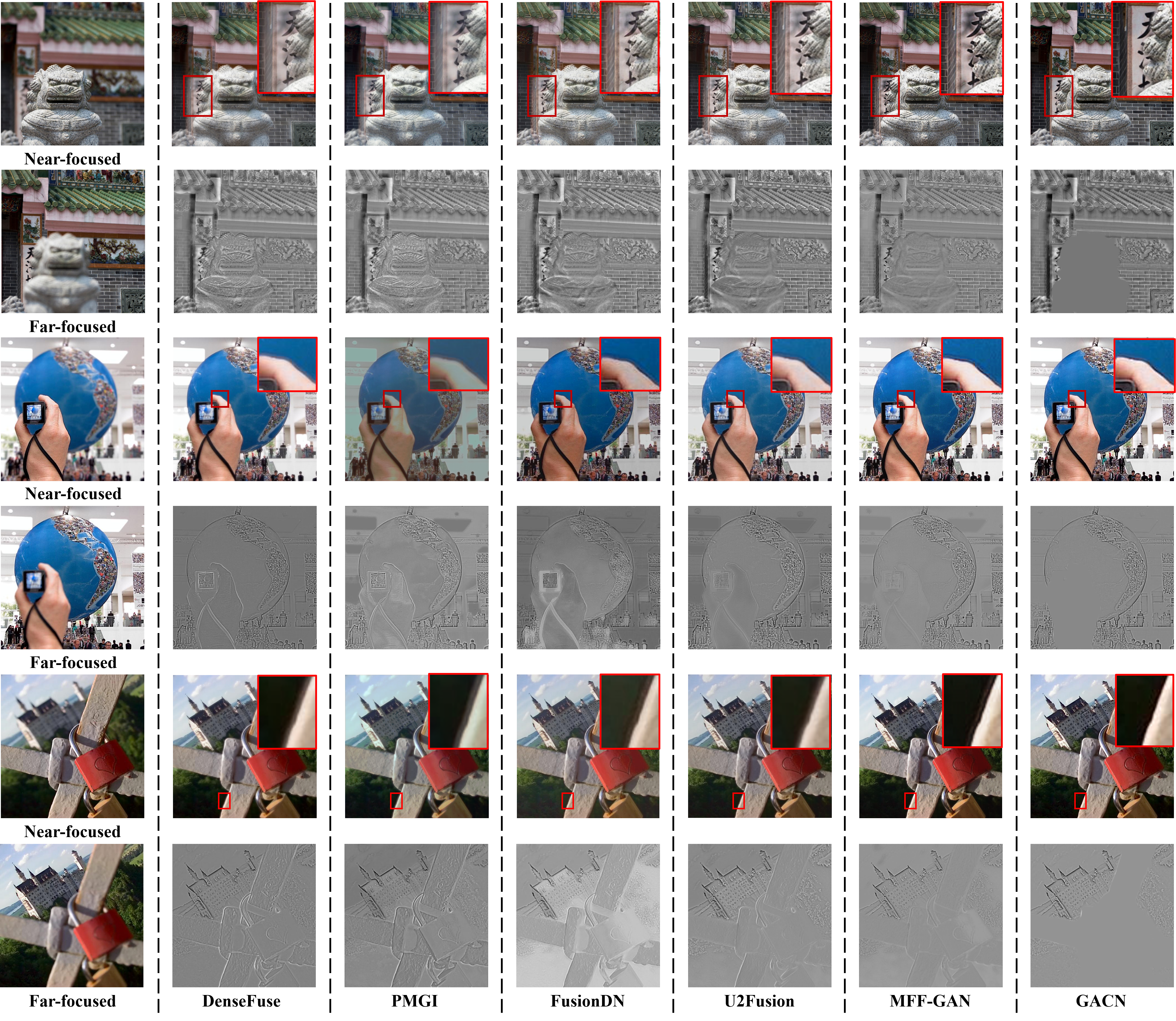} 
		\caption{Visualization of fusion result and difference images of decoder based methods (Dense-Fuse, PMGI, FusionDN, U2Fusion, and MFF-GAN) and GACN. For each example, the top image is fusion result and the bottom image is difference image, which is obtained by subtracting the near-focused image from the fusion result.}
		\label{fig:vis_compare_decoder}
	\end{figure*}

	\subsubsection{Subjective Assessment}
	\label{sec:subjective_assessments}
	
	We show some visualization results of GACN and other SOTA methods, DM based and decoder based methods, respectively. Firstly, we present the decision maps of GACN with some classical DM based methods (CNN-Fuse and SESF-Fuse) in Figure~\ref{fig:vis_compare_dm}. The influence of post processing method is shown in detail. According to the experiment, the SESF-Fuse and CNN-Fuse require post-processing methods with empirical parameters, such as morphology operation and small size removal strategy, to eliminate noise. If we remove these post processing methods, there will be some artifacts that appear on the results, such as blob noisy in the decision map. Besides, the threshold of kernel size in morphology operation and region removal strategy are empirical parameters which hard to adjust. While our method GACN can draw out good decision map without post-processing methods. 
	
	Secondly, we demonstrate the fusion results and difference images of GACN with some classical decoder based methods (Dense-Fuse, PMGI, FusionDN, U2Fusion and MFF-GAN) in Figure~\ref{fig:vis_compare_decoder}. The red rectangles and their magnified regions (shown in upper right of the figure) denote the detailed fusion results of different methods. It is shown that there is artifact area at the border of near-focused and far-focused regions for the classical decoder based methods. While GACN shows clear result. The difference image is obtained by subtracting the near-focused image from the fusion result, which is normalized to the range of 0 to 1 for visualization. If the near-focused region is completely detected, the difference image will not show any of its information. Decoder based methods cannot precisely recover the true pixel values in fusion result due to the nonlinear mapping mechanism in the decoder. Therefore most of them have clear contour information in the near-focused region on the difference images. Besides, there is some color distortion in the fusion result of PMGI. And the fusion result of DenseFuse is more blurred than other methods. Fortunately, our method, GACN, achieves robust promising fusing performance on all examples.

	\subsection{Ablation Study}
	\label{sec:ablation_study}
	We evaluate our method with different settings to verify the contribution of each module. 
	
	\subsubsection{Loss Function Study}
\begin{figure}[ht]
		\centering
		\includegraphics[width=0.6\linewidth]{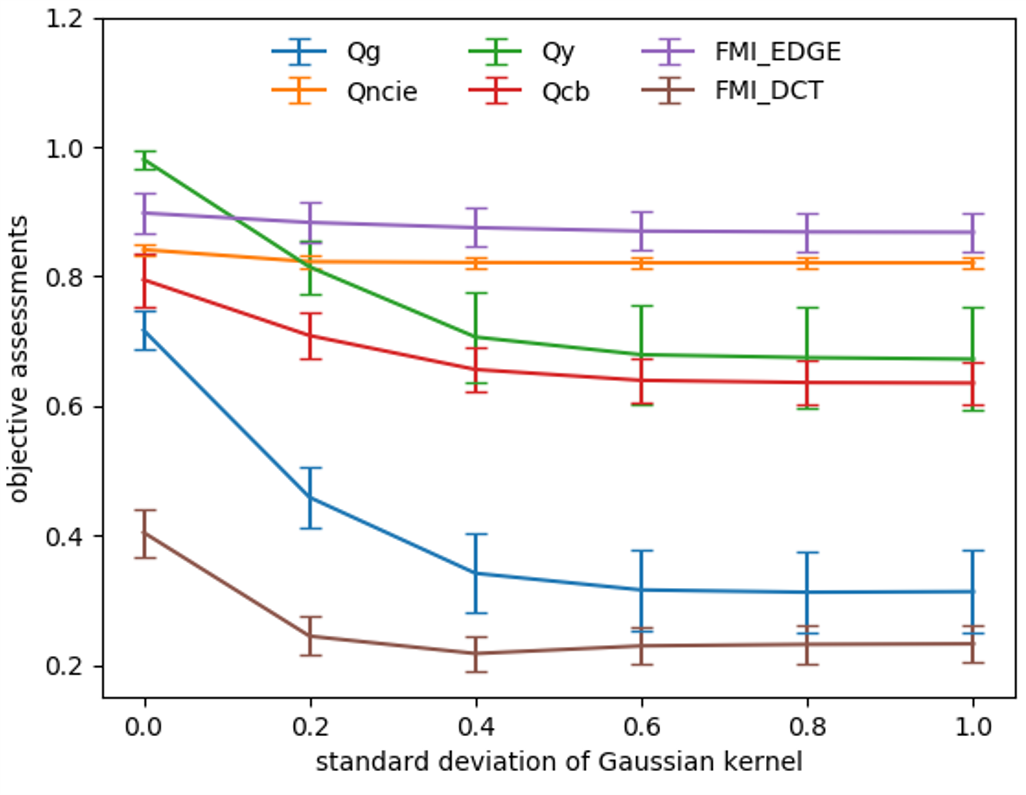}
		\caption{Variation of different metrics.}
		\label{fig:diff_variant_metrics}
	\end{figure}

	We first conducted an experiment to figure out which metric is more suitable for evaluation of quality of multi-focus image fusion. We introduce Gaussian blurring with different standard deviations to the fusion result of the testing set. As shown in Figure~\ref{fig:diff_variant_metrics}, with the increase of standard deviation of Gaussian kernel, the metric $Q_g$ degenerates most obviously compared to other metrics. It is shown that the metric $Q_g$ can better reflect the clarity of the fusion result, which means that metric $Q_g$ is beneficial to be the loss function for model training.
	
\begin{figure}[ht]
		\centering
		\includegraphics[width=1.0\linewidth]{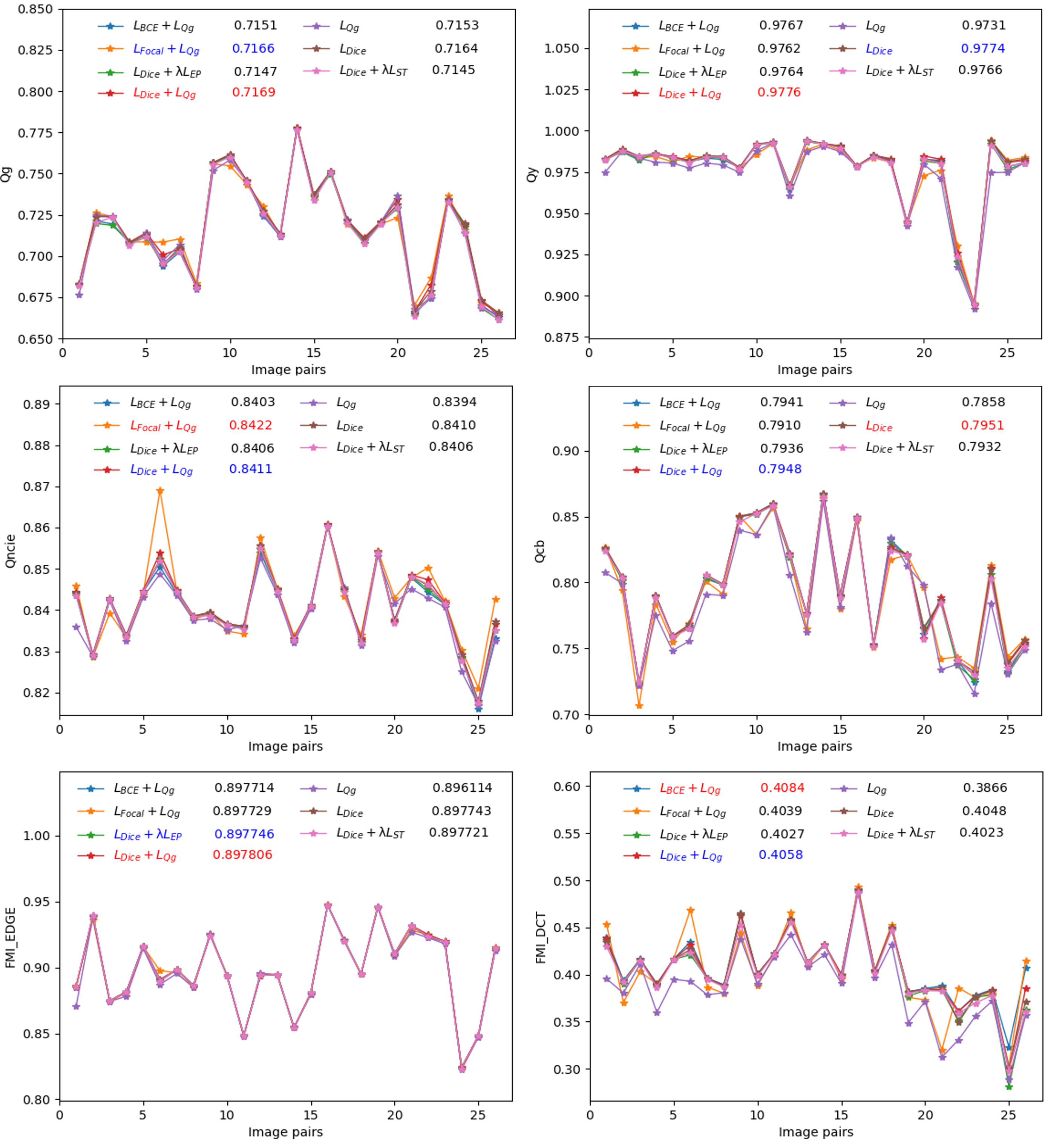}
		\caption{Loss Function Analysis.}
		\label{fig:loss_study}
	\end{figure}

	In addition, we compared the performance of the different combinations of mask-based and gradient-based loss functions to verify the contribution of proposed loss functions, shown in Figure~\ref{fig:loss_study}. The mask based loss functions include $L_{Dice}$~\cite{milletari2016vnet}, $L_{Focal}$~\cite{lin2017focal}, and $L_{BCE}$~\cite{xie2015holistically}. While gradient-based loss functions include $L_{Q_g}$, $L_{EG}$~\cite{li2020drpl}, and $L_{ST}$~\cite{Jung2020U}. $L_{BCE}$ denotes balanced cross entropy which is a classical loss function in image segmentation~\cite{xie2015holistically}, which can eliminate the impact of imbalance pixels in the foreground and background. $L_{Focal}$ denotes focal loss~\cite{lin2017focal}, which leads the network to focus on and correctly detect hard examples. Where $EG$ refers to edge-preserving loss and ST means structure tensor loss. For the last two losses, we conducted an experiment and pick the best performance with $\lambda = 0.0001$ to balance the importance with $L_{Dice}$. According to the experiment, we noted that the performance of the combination of $L_{Dice}$ and $L_{Q_g}$ outperforms other loss settings in most the metrics, which means that the above two losses will both lead the network to export promising fusing result. Besides, we find that $L_{Dice}$ is better than $L_{BCE}$, and $L_{Focal}$, which means that $L_{Dice}$ can precisely recognize the decision map. And $L_{Q_g}$ is better than $L_{EP}$, and $L_{ST}$, which means that $L_{Q_g}$ can better lead the structure to preserve gradient information in the fused result.
	
	\begin{figure*}[!t]
		\centering
		\includegraphics[width=1.0\linewidth]{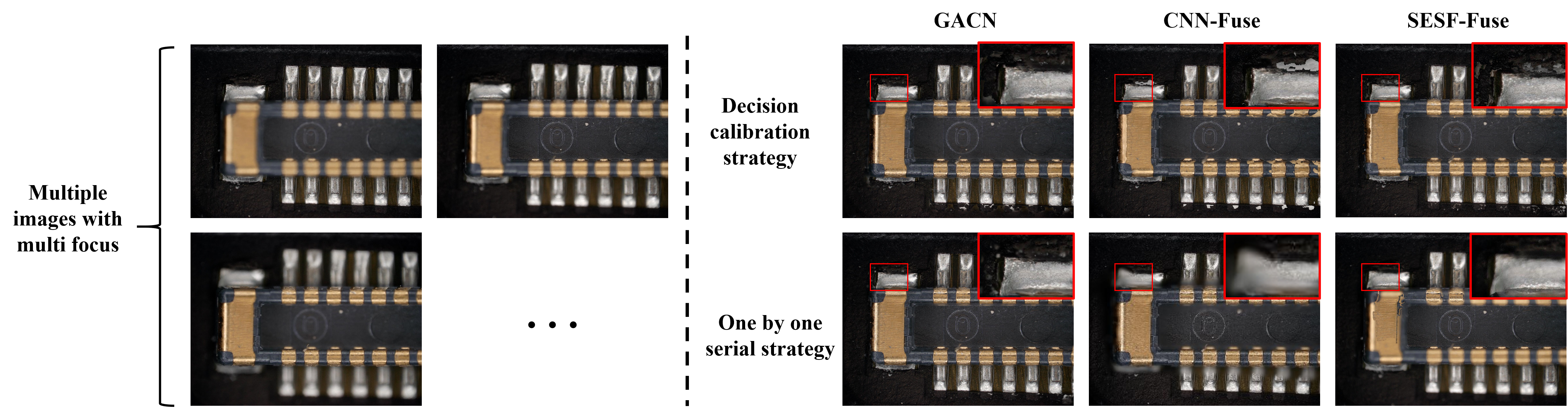}
		\caption{Visualization of multiple images fusion.}
		\label{fig:multi_images_fusion}
	\end{figure*}
	
	\subsubsection{Differentiation Study}
	We compared the performance of the absolute function and the smooth approximation for angle calculation (Eq.~\ref{eq:loss_qg_4}) in $L_{Q_g}$ in Table~\ref{table:differentiation_study}. We found that directly using the absolute function is a little better than the smooth approximation by using pytorch framework, which might be the reason for the gradient vanish in the sigmoid calculation.
	
	\begin{table}[ht]
		\centering
		\caption{Differentiation Comparison. 'Abs' means absolute function, and 'Smooth' denotes smooth approximation.}
		\label{table:differentiation_study}
		\begin{tabular}{cccc}
			\hline
			Settings & $Q_g$           & $Q_y$                         & $Q_{nice}$      \\ \hline
			Abs      & \textbf{0.7169} & \textbf{0.9776}               & \textbf{0.8411} \\
			Smooth   & 0.7162          & 0.9773                        & 0.8410          \\ \hline
			Settings & $Q_{cb}$        & $FMI\_EDGE$                   & $FMI\_DCT$      \\ \hline
			Abs      & 0.7948          & \textbf{0.8978}               & \textbf{0.4058} \\
			Smooth   & \textbf{0.7952} & 0.8977                        & 0.4048          \\ \hline
		\end{tabular}
	\end{table}

	\subsection{Multiple images fusion with multi-focus}
	\label{sec:results_on_multi_images_fusion}
	The example of multiple images fusion is shown in Table~\ref{table:time_consumption_of_multiple_images} and Figure \ref{fig:multi_images_fusion}. The microscopic image 'chip' (with the size of $2700 \times 1800$) was obtained by a microscope that took pictures with lots of different focus points. Decision calibration for 'chip' images fusion can actually increase execution efficiency by about 30.65\% compared to one-by-one serial strategy (0.7138's to 0.4905's for each image by using GACN), which is more feasible for industrial application. And the same increase of efficiency can also be found in CNN-Fuse and SESF-Fuse, which means that the decision calibration can be applied to other networks. Note that for decision calibration, we deleted the post-processing operations of DM in CNN-Fuse and SESF-Fuse for fair comparison. 
	
	The visualization of fusion result of GACN is more clear than that of CNN-Fuse and SESF-Fuse whether in decision calibration or serial strategy. In the future work, we try to eliminate the impact of defocused spread effect for multi-focus image fusion.
	
	
	\begin{table}[ht]
		\centering
		\caption{Time consumption per image for multiple 'chip' images fusion with multi-focus points. The bold value denotes the best performance in each method. CNN-Fuse is running on CPU mode according to its public code.}
		\label{table:time_consumption_of_multiple_images}
		\begin{tabular}{ccc}
			\hline
			Runtime(s)      & One by one serial             & Decision calibration  \\ \hline
			CNN-Fuse        & 886.6872                      & \textbf{687.5352}      \\
			SESF-Fuse       & 1.1880                        & \textbf{0.5293}        \\
			GACN            & 0.7138                        & \textbf{0.4905}        \\ \hline
		\end{tabular}
	\end{table}

	
	\section{Conclusion}
	\label{sec:conclusion}
	In this work, we propose a network to simultaneously generate decision map and fused result with an end-to-end training procedure. It avoids utilizing empirical post-processing methods in the inference stage. Besides we introduce a gradient aware loss function to lead the network to preserve gradient information. Also we design a decision calibration strategy to fuse multiple images, which can increase implementation efficiency. Extensive experiments are conducted to compare with existing SOTA multi-focus image fusion structures, which shows that our designed structure can generally ameliorate the output fused image quality for multi-focus images, and increase implementation efficiency over 30\% for multiple images fusion. We will further improve the fusion performance of multiple images fusion in future work.
	
	\section{Acknowledgments}
	\label{sec:acknowledgments}
	The authors would like to thank professor Jiayi Ma of Wuhan University for the advice about the visualization of subjective assessment. In addition, we thank Zhuhai Boming Vision Technology Co., Ltd by providing the dataset of multiple images fusion. The computing work is supported by USTB MatCom of Beijing Advanced Innovation Center for Materials Genome Engineering.
	
	\bibliography{refs}
\end{document}